\documentclass{article}

\usepackage[preprint,nonatbib]{neurips_2024}

\usepackage[utf8]{inputenc}     % allow utf-8 input
\usepackage[T1]{fontenc}        % use 8-bit T1 fonts
\usepackage{url}                % simple URL typesetting
\usepackage{booktabs}           % professional-quality tables
\usepackage{multirow}           % for multirows in tables
\usepackage{adjustbox}
\usepackage{amsmath,amssymb}    % additional characters
\usepackage{bm}                 % bold math symbols
\usepackage{amsfonts}           % blackboard math symbols
\usepackage{nicefrac}           % compact symbols for 1/2, etc.
\usepackage{microtype}          % microtypography
\usepackage{stmaryrd}           % for \llbracket
\usepackage{subcaption}         % for subfigures
\usepackage[dvipsnames]{xcolor} % colors

\definecolor{tyellow1}{HTML}{FCE94F}
\definecolor{tyellow2}{HTML}{EDD400}
\definecolor{tyellow3}{HTML}{C4A000}
\definecolor{torange1}{HTML}{FCAF3E}
\definecolor{torange2}{HTML}{F57900}
\definecolor{torange3}{HTML}{C35C00}
\definecolor{tbrown1}{HTML}{E9B96E}
\definecolor{tbrown2}{HTML}{C17D11}
\definecolor{tbrown3}{HTML}{8F5902}
\definecolor{tgreen1}{HTML}{8AE234}
\definecolor{tgreen2}{HTML}{73D216}
\definecolor{tgreen3}{HTML}{4E9A06}
\definecolor{tblue1}{HTML}{729FCF}
\definecolor{tblue2}{HTML}{3465A4}
\definecolor{tblue3}{HTML}{204A87}
\definecolor{tpurple1}{HTML}{AD7FA8}
\definecolor{tpurple2}{HTML}{75507B}
\definecolor{tpurple3}{HTML}{5C3566}
\definecolor{tred1}{HTML}{EF2929}
\definecolor{tred2}{HTML}{CC0000}
\definecolor{tred3}{HTML}{A40000}
\definecolor{tlgray1}{HTML}{EEEEEC}
\definecolor{tlgray2}{HTML}{D3D7CF}
\definecolor{tlgray3}{HTML}{BABDB6}
\definecolor{tdgray1}{HTML}{888A85}
\definecolor{tdgray2}{HTML}{555753}
\definecolor{tdgray3}{HTML}{2E3436}

\usepackage[
 colorlinks=true    % false: boxed links; true: colored links
,linkcolor=tblue3   % color of internal links
,citecolor=tblue3   % color of links to bibliography
,filecolor=black    % color of file links
,urlcolor=tblue3    % color of external links
,linktoc=page       % only page is linked
,plainpages=false]{hyperref}

\usepackage{cleveref}

\usepackage{tikz}
\usetikzlibrary{arrows}
\usetikzlibrary{automata}
\usetikzlibrary{backgrounds}
\usetikzlibrary{calc}
\usetikzlibrary{chains}
\usetikzlibrary{circuits.logic.US}
\usetikzlibrary{decorations.footprints}
\usetikzlibrary{decorations.markings}
\usetikzlibrary{decorations.pathmorphing}
\usetikzlibrary{decorations.text}
\usetikzlibrary{fadings}
\usetikzlibrary{fit}
\usetikzlibrary{graphs}
\usetikzlibrary{matrix}
\usetikzlibrary{mindmap}
\usetikzlibrary{overlay-beamer-styles}
\usetikzlibrary{patterns}
\usetikzlibrary{positioning}
\usetikzlibrary{scopes}
\usetikzlibrary{shadows}
\usetikzlibrary{shapes.arrows}
\usetikzlibrary{shapes.geometric}
\usetikzlibrary{shapes.misc}
\usetikzlibrary{shapes.symbols}
\usetikzlibrary{shapes}
\usetikzlibrary{snakes}
\usetikzlibrary{trees}

\graphicspath{{figures/}}

\usepackage[
  sortcites,
  style=numeric,
  sorting=nyt]{biblatex}
\title{Formal Explanations for Neuro-Symbolic AI}

\newcommand{\mailtodomain}[1]{\href{mailto:#1@monash.edu}{\nolinkurl{#1}}}

\author{%
  \hspace{-0.5cm}Sushmita Paul \hspace{0.2cm} Jinqiang Yu \hspace{0.2cm} Jip J. Dekker \hspace{0.2cm} Alexey Ignatiev \hspace{0.2cm} Peter J. Stuckey \\
  \hspace{-0.5cm}Department of Data Science and AI, Faculty of IT\\
  \hspace{-0.5cm}Monash University, Melbourne, Victoria, Australia \\
  \hspace{-0.5cm}\{\mailtodomain{sushmita.paul}\texttt{,}\mailtodomain{jinqiang.yu}\texttt{,}\mailtodomain{jip.dekker}\texttt{,}\mailtodomain{alexey.ignatiev}\texttt{,}\mailtodomain{peter.stuckey}\}\texttt{@monash.edu} \\
}

\newtheorem{proposition}{Proposition}

\newtheorem{definition}{Definition}

\newtheorem{example}{Example}

\newcommand{\fml}[1]{{\mathcal{#1}}}

\newcommand{\mbf}[1]{\ensuremath\mathbf{#1}}
\newcommand{\mbb}[1]{\ensuremath\mathbb{#1}}

\newcommand{\ignore}[1]{}

\DeclareMathOperator*{\entails}{\vDash}
\DeclareMathOperator*{\nentails}{\nvDash}

\DeclareMathOperator*{\limply}{\rightarrow}

\def\squareforqed{\hbox{\rlap{$\sqcap$}$\sqcup$}}
\def\qed{\ifmmode\squareforqed\else{\unskip\nobreak\hfil
\penalty50\hskip1em\null\nobreak\hfil\squareforqed
\parfillskip=0pt\finalhyphendemerits=0\endgraf}\fi}

\begin{document}

\maketitle

\begin{abstract}
  Despite the practical success of Artificial Intelligence (AI), current 
  neural AI algorithms face two significant issues.
  First, the decisions made by neural architectures are often
  prone to bias and brittleness. Second, when a chain of reasoning is
  required, neural systems often perform poorly.
  Neuro-symbolic
  artificial intelligence is a promising approach
  that tackles these (and other) weaknesses by combining the 
  power of neural perception and symbolic reasoning.
  Meanwhile, the success of AI has made it
  critical to understand its behaviour,
  leading to the development of
  explainable artificial intelligence (XAI).
  While neuro-symbolic AI systems have important advantages over
  purely neural AI, we still need to explain their
  actions, which are obscured by the interactions of the neural and
  symbolic components.
  To address the issue, this paper proposes a formal approach to
  explaining the decisions of neuro-symbolic systems.
  The approach hinges on the use of formal abductive explanations and
  on solving the neuro-symbolic explainability problem hierarchically.
  Namely, it first computes a formal explanation for the symbolic
  component of the system, which serves to identify a subset of the
  individual parts of neural information that needs to be explained.
  This is followed by explaining only those individual neural inputs,
  independently of each other, which facilitates succinctness of
  hierarchical formal explanations and helps to increase the overall
  performance of the approach.
  Experimental results for a few complex reasoning tasks
  demonstrate practical efficiency of the proposed approach, in
  comparison to purely neural systems, from the perspective of
  explanation size, explanation time, training time, model sizes, and the quality of explanations reported.
\end{abstract}

\section{Introduction} \label{sec:intro}

Neural Artificial Intelligence (AI) models are widely used
by  the decision-making procedures of many real-world applications.
Their success guarantees AI will prevail as a generic
computing paradigm for the foreseeable future, including in safety-
and privacy-critical domains.
Unfortunately, neural AI models may occasionally
fail~\cite{theguardian16,cnn18,forbes18},
their decisions may be prone to bias~\cite{propublica16} or be
confusing due to brittleness~\cite{szegedy-iclr14,goodfellow-iclr15},
where very similar cases are treated completely differently.
As a result, there is a need to understand the behaviour of AI models, analyse their
potential failures, debug them, and possibly repair
them. This has given rise to AI model
verification~\cite{kwiatkowska-ijcai18,narodytska-aaai18,lomuscio-kr18,barrett-cav19}
and explainable AI
% (XAI)~\cite{guestrin-kdd16,guestrin-aaai18,lundberg-nips17,darwiche-ijcai18,inms-aaai19,monroe-cacm18,lipton-cacm18,miller-aij19,msi-aaai22,yu-aaai23,barrett-neurips23}.
(XAI)~\cite{monroe-cacm18,lipton-cacm18,miller-aij19}.

Many XAI approaches exist, including
the computation of interpretable AI
models~\cite{leskovec-kdd16,rudin-kdd17a,molnar-bk20} and post-hoc
explanation extraction for black-box AI
models~\cite{guestrin-kdd16,lundberg-nips17,guestrin-aaai18}.
An important approach to XAI -- \emph{formal explainable AI} (FXAI) -- builds on the use of
formal reasoning about the AI models of interest and on computing explanations that
capture the semantics of the target models to answer \emph{``why''}
or \emph{``why not''} they make certain
decisions~\cite{darwiche-ijcai18,inms-aaai19,msi-aaai22}.

Neuro-symbolic AI is a promising paradigm developed
to add symbolic reasoning to neural AI systems~\cite{dillig-neurips20,deraedt-aij21,ahmed-aaai22,russo-ml22,naik-pldi23}
by  combining neural components (e.g., for sensory
understanding) with symbolic ones (e.g., for reasoning about sensory
results).
Neuro-symbolic AI systems can significantly outperform both purely neural and purely
symbolic systems -- as shown in several practical
domains~\cite{deraedt-aij21,ahmed-aaai22, naik-pldi23} -- by taking advantage of the strengths of
the two approaches, and are deemed vital for constructing rich
computational cognitive
models~\cite{valiant-fsttcs08,marcus-corr20,garcez-air23}.
However,
%despite showing promise in
%several practical
%domains~\cite{deraedt-aij21,ahmed-aaai22,naik-pldi23}, they inherit
%some potential failures, decision bias, and
%brittleness intrinsic to neural AI.
%
since their neural component is still an opaque black-box, the resulting
decisions can still be hard or impossible for humans
to comprehend.
Moreover, interactions between the neural and symbolic
components of the system exacerbate these problems, further
deteriorating human understanding of the system's operation.

This paper addresses these challenges by building on FXAI~\cite{msi-aaai22}
to propose the first formal approach to the explainability of modern
neuro-symbolic systems.
In doing so, it considers the extraction of abductive explanations from neuro-symbolic systems
where neural perception inputs are passed to the symbolic reasoner,
independently of each other.
The problem is tackled hierarchically, that is,
given a decision made by a complex neuro-symbolic system, we
start by explaining the decision of the latest component of the system
and identify a subset of its inputs (passed through by the earlier
components) responsible for the decision.
%
%Namely, given a decision made by a complex neuro-symbolic system, we
%start by explaining the decision of the latest component of the system
%and identify a subset of its inputs (passed through by the earlier
%components) responsible for the decision.
%
The approach then proceeds by explaining why the latest component
received these particular inputs, going backwards through the
neuro-symbolic interaction.
Our experimental results on a range of problem families
demonstrate the usefulness and practicality of the proposed approach and its advantage
over the baseline (purely neural) approach in terms of explanation
size and quality. This in turn serves as an additional motivation for the
development of next-generation neuro-symbolic AI.

The paper is organized as follows.
\autoref{sec:prelim} overviews the background information and introduces 
the required concepts.
\autoref{sec:app} discusses the proposed approach followed by the
experimental results in \autoref{sec:res}.
\autoref{sec:relw} briefly outlines related work and
\autoref{sec:conc} concludes the paper.

% \anote{Here we need to motivate the problem:
% \begin{itemize}
%   \item current generation of ML is not without issues and requires XAI
%   \item Heuristic vs formal XAI: pros and cons
%   \item Neuro-symbolic AI seeks to address the weaknesses of neural ML
%   \item This paper is about... Contributions.
%   \item Paper outline
% \end{itemize}
% }

\section{Preliminaries} \label{sec:prelim}

% This section provides a brief overview of the concepts and notations
% used throughout the paper.

\paragraph{Datalog, Satisfiability and Unsatisfiability.}
Datalog is a declarative language defined as a subset of the logic programming language Prolog, with a
focus on data manipulation and querying~\cite{abiteboul-bk95}.
In particular, its limited
features enable the use of bottom-up evaluation, rather than the top-down
evaluation common in Prolog.
This makes it well-suited as a query language for
deductive databases, where it can infer new facts based on existing
data and rules.
Answer Set Programming (ASP) languages~\cite{niemela-aimag16} extend Datalog
to include constraints and use propositional satisfiability (SAT) solving methods.
\ignore{
Datalog also is a syntactic subset of Answer Set Programming (ASP)
languages~\cite{niemela-aimag16},
which focus on solving complex decision problems, using
propositional satisfiability (SAT) solving methods.
While ASP offers greater expressive power, Datalog's simplicity and
guaranteed termination make it a valuable tool for specific tasks
within the logic programming domain.
}

We assume standard definitions for SAT and maximum satisfiability
(MaxSAT) solving~\cite{sat-handbook}.
A propositional formula is said to be in \emph{conjunctive}
\emph{normal form} (CNF) if it is a conjunction of clauses, where a
\emph{clause} is a disjunction of literals, and a
\emph{literal} is either a Boolean variable or its negation.
Whenever convenient, clauses are treated as sets of literals while CNF
formulas are treated as sets of clauses.
A \emph{truth assignment} maps each variable of a formula to a value from
$\{0,1\}$.
Given a truth assignment, a clause is said to be \emph{satisfied} if at least
one of its literals is assigned value 1; otherwise, it is \emph{falsified} by
the assignment.
A formula $\phi$ is satisfied if all its clauses are satisfied; otherwise,
it is falsified.
If there exists no assignment that satisfies $\phi$, then it
is \emph{unsatisfiable}, written $\phi \models \bot$.

In the context of unsatisfiable formulas, the problems of maximum
satisfiability and minimum unsatisfiability are of particular
interest.
%
% to find a truth assignment that maximizes the number of satisfied
% clauses.
%
% While a number of variants of MaxSAT
% exist~\cite[Ch.~23~and~24]{sat-handbook}, hereinafter, we are
% interested in Partial Unweighted MaxSAT, which can be formulated as
% follows.
%
Hereinafter, we consider \emph{partial} unsatisfiable CNF formulas
$\phi$ represented as a conjunction of \emph{hard} clauses $\fml{H}$,
which must be satisfied, and \emph{soft} clauses $\fml{S}$, which are
preferred to be satisfied, i.e.,\ $\phi=\fml{H}\land\fml{S}$ (or
$\phi=\fml{H}\cup\fml{S}$ in the set theory notation).
%
% The Partial Unweighted MaxSAT problem consists in finding an
% assignment that satisfies all hard clauses and maximizes the total
% number of satisfied soft clauses.
%
In the analysis of unsatisfiability of formula $\phi$, one is
interested in identifying minimal unsatisfiable subsets (MUSes) of
$\phi$, which can be defined as follows.
Let $\phi=\fml{H}\cup\fml{S}$ be unsatisfiable, i.e.,\
$\phi\entails\bot$.
A subset of clauses $\mu\subseteq\fml{S}$ is a {\em  Minimal
Unsatisfiable Subset} (MUS) iff $\fml{H}\cup\mu\entails\bot$ and
$\forall{\mu'\subsetneq\mu}$ it holds that
$\fml{H}\cup\mu'\nentails\bot$.
Informally, an MUS can be seen as a minimal explanation of
unsatisfiability for an unsatisfiable formula $\phi$.
It provides the minimal information that needs to be added to the hard
clauses $\fml{H}$ to obtain unsatisfiability.
MUS extraction is applied in our work as the underlying technology for
computing formal explanations for symbolic and neural AI systems.
% \pjs{Do we ever use SAT, MaxSAT. We do use MUS?} \anote{Yes, we use
% MUSes. To explain MUS/unsatisfiability, we define SAT.}

\paragraph{Classification Problems}
classify data instances into classes
$\fml{K}$ where $|\fml{K}| = k \geq 2$.
Given a set of $m$ features $\fml{F}$, where the value of
feature $i \in \fml{F}$ comes from a domain $\mbb{D}_i$ which may be
Boolean, (bounded) integer or (bounded) real, the \emph{complete feature space} is defined by
$\mathbb{F}\triangleq\prod_{i=1}^{m}\mbb{D}_i$.
A \emph{data point} in feature space is denoted $\mbf{v} = (v_1,
\ldots, v_m)$ where $v_i \in \mbb{D}_i, 1 \leq i \leq m$.
An \emph{instance} of a classification problem is a pair formed by a data point and its corresponding class, i.e.,
$(\mbf{v}, c)$, where $\mbf{v}\in\mbb{F}$ and $c\in \fml{K}$.
We use $\mbf{x} = (x_1, \ldots, x_m)$ to denote an
arbitrary point in feature space, where each $x_i$ will take a value
from $\mbb{D}_i$.
A \emph{classifier} is a total function from feature space to class:
$\kappa: \mathbb{F} \rightarrow \fml{K}$.
Many approaches exist to define classifiers including decision
sets~\cite{clark-ewsl91,leskovec-kdd16}, decision
lists~\cite{rivest-ml87}, decision trees~\cite{rivest-ipl76}, random
forests~\cite{friedman-tas01}, boosted trees~\cite{guestrin-kdd16a},
and neural nets~\cite{hinton-icml10,hcseyb-neurips16}.

\begin{figure}[t]
  \centering
  \begin{subfigure}[b]{0.24\textwidth}
    \centering
    \includegraphics[height=3.3cm]{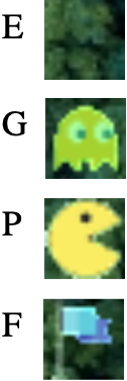}
    \caption{Image classification}
    \label{fig:class}
  \end{subfigure}%
  \hfill
  \begin{subfigure}[b]{0.24\textwidth}
    \includegraphics[height=3.3cm]{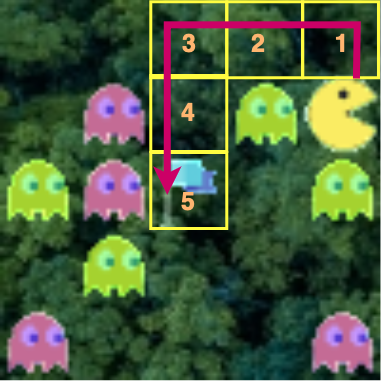}
    \caption{Input problem instance}
    \label{fig:grid}
  \end{subfigure}%
  \hfill
  \begin{subfigure}[b]{0.24\textwidth}
    \includegraphics[height=3.3cm]{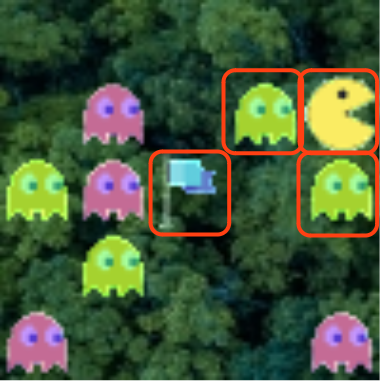}
    \caption{Symbolic explanation}
    \label{fig:expl}
  \end{subfigure}
  \hfill
  \begin{subfigure}[b]{0.24\textwidth}
    \includegraphics[height=3.3cm]{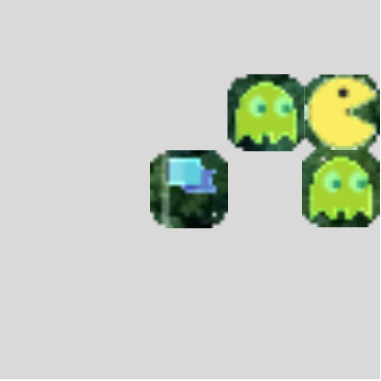}
    \caption{Neural explanation}
    \label{fig:nexpl}
  \end{subfigure}
  \caption{Example of a Pacman related puzzle aiming to find a
  shortest grid path from the actor's location to the target flag.
  Given an input instance, the model predicts the length of a
  shortest path.}
  \label{fig:pacman}
\end{figure}

%\paragraph{Running Example.}
%%
%To illustrate the ideas behind neuro-symbolic systems and their
%explanations, the following running example and its variations
%from~\cite{li-aaai23} will be used throughout the paper.
%%
\begin{example}
  Consider a simple classification function that takes a $\text{20}
  \times \text{20}$ RGB image $\mbf{x}\in\mbb{F}$ (original data are
  taken from~\cite{naik-pldi23}) and decides if it represents an empty
  cell $E$, a ghost $G$, an actor $P$, or a flag $F$.
  Example images are shown in \autoref{fig:class}.
  We can train a neural net classifier $\kappa: \mbb{F} \rightarrow \{
  E,G,P,F\}$ to be very accurate.
  \qed
\end{example}

\paragraph{Neuro-Symbolic AI}
is an important paradigm that bridges
the gap between neural systems and logic and reasoning
systems~\cite{dillig-neurips20,deraedt-aij21,ahmed-aaai22,russo-ml22,naik-pldi23}.
It combines the strengths of both symbolic and subsymbolic approaches
by integrating logical reasoning with the powerful capabilities of
deep learning.
This allows neuro-symbolic models to leverage explicit domain
knowledge and data-driven insights, leading to more accurate and
efficient solutions for challenging AI
tasks~\cite{valiant-fsttcs08,marcus-corr20,garcez-air23}.
Scallop~\cite{naik-pldi23}
is neuro-symbolic framework that integrates symbolic reasoning
capabilities, exposed through its extended Datalog language, with the
power of deep learning provided by PyTorch~\cite{ansel-asplos24}.
%
%Its key strengths  lie in the capabilities of the
%Scallop programming language.
%
Through its support for differentiable programming, Scallop can seamlessly
use tensors of neural networks as input or output for a symbolic
program.
Conversely, the same capabilities allow the back-propagation for the
neural networks through a Scallop program using predefined
probabilistic semantics of \emph{provenance semirings}~\cite{green-pods07}.

\begin{example} \label{ex:rex}
  Consider a Pacman related puzzle with the task to find the length of
  the shortest grid path from the actor's start location to a target
  flag while avoiding ghosts.
  The input to the puzzle shown in \autoref{fig:grid} is an image
  consisting of $\text{5} \times \text{5}$ grid cells with background
  and icons of either actor, target, or ghosts imposed on top.
  The input is a $\text{100} \times \text{100}$ RGB image, and the
  output is the length of the shortest path from the unique actor
  location to the unique flag location avoiding entering any ghost
  location.
  Observe how \autoref{fig:grid} also depicts a shortest path, shown
  in red.
  \qed
\end{example}

% \pjs{Sushmita, add four little squares to the left of the existing figure
%   with examples of the four classes. Thanks
% }

\paragraph{Formal Explainability and Adversarial Robustness.}
Given data point $\mbf{v}$ and classifier $\kappa$, which classifies it as
class $\kappa(\mbf{v})$,
a \emph{post-hoc explanation} tries to explain the behaviour of $\kappa$ on $\mbf{v}$.
%
% \ignore{
% We consider two forms of formal explanation answering \emph{why} and
% \emph{why not} (or \emph{how}) questions.
% }
%
An \emph{abductive explanation} (AXp) is a minimal set of features
$\fml{X}$ such that any data point sharing the same feature values
with $\mbf{v}$ on these features is guaranteed to be assigned the same
class $c = \kappa(\mbf{v})$~\cite{darwiche-ijcai18,inms-aaai19}.
Formally, $\fml{X}$ is a subset-minimal set of features such that:
\begin{equation} \label{eq:axp}
  \forall(\mbf{x} \in \mbb{F}). \left[\bigwedge\nolimits_{i \in \fml{X}}
  (x_i = v_i)\right] \limply (\kappa(\mbf{x}) = c)
\end{equation}

%
%A \emph{contrastive explanation} (CXp) for the classification of data
%point $\mbf{v}$ with class $c = \kappa(\mbf{v})$ is a minimal set of
%features that must change so that $\kappa$ can return a different
%class.
%%
%Formally, a CXp is a subset minimal set of features $\fml{Y}$ such
%that
%%
%\begin{equation} \label{eq:cxp} %
%\exists(\mbf{x}\in\mbb{F}).\left[\bigwedge\nolimits_{i\not\in\fml{Y}}(x_i=v_i)\right]\land(\kappa(\mbf{x})\not=c)
%% \end{equation}

It is known~\cite{inams-aiia20,msi-aaai22} that formal AXps for ML predictions
are related with the concept of MUSes (defined earlier) of an
\emph{unsatisfiable} formula encoding the ML classification process
$\kappa(\mbf{v})=c$, namely if one represents
$\left[\kappa(\mbf{x})\not=c\right]$ as hard clauses and
$\left[\bigwedge_{i=1}^m{(x_i=v_i)}\right]$ as soft clauses.
%%
%For this reason, the set $\axps$ of all AXp's $\fml{X}$ explaining
%classification $\kappa(\mbf{v}) = c$ and the set $\cxps$ of all CXp's
%$\fml{Y}$ explaining the same classification enjoy a \emph{minimal
%hitting set duality}~\cite{inams-aiia20}, similarly to MUSes and
%MCSes.
%%
%That is $\axps = \mhs(\cxps)$ and is $\cxps = \mhs(\axps)$.
%%
%This property can be made use of when computing or enumerating AXp's
%and/or CXp's~\cite{inams-aiia20,msgcin-icml21,msi-aaai22}.

By examining \eqref{eq:axp}, one can observe that AXps are designed to
hold globally, i.e.,\ in the entire feature space $\mbb{F}$.
Recent work proposed to use the tools developed for checking
adversarial robustness of neural networks in the context of computing
\emph{distance-restricted} abductive
explanations~\cite{barrett-neurips23,huang-corr23b}.
Namely, instead of requiring the explanation $\fml{X}$ to hold for all
the points $\mbf{x}\in\mbb{F}$ s.t.\
$\bigwedge_{i\in\fml{X}}{(x_i=v_i)}$, we can enforce it for
all compatible points in the $\varepsilon$-vicinity of the instance
$\mbf{v}$ of interest:
\begin{equation} \label{eq:eaxp}
  \forall(\mbf{x} \in \mbb{F}).
  \left[
    \bigwedge\nolimits_{i \in \fml{X}} (x_i = v_i)\land
    \lVert\mbf{x}-\mbf{v}\rVert_p\leq\varepsilon
  \right]
  \limply (\kappa(\mbf{x}) = c)
\end{equation}
where the $\varepsilon$-neighbourhood, s.t. $\varepsilon\in[0,1]$, is
defined given some $p$-norm,
$p\in\{0,1,\ldots,\infty\}$~\cite{horn-bk12,robinson-bk03}.
When $\varepsilon=1$, predicate \eqref{eq:eaxp} equates with
\eqref{eq:axp}.
When formally explaining a neural model, this work makes use of an
adversarial robustness checker~\cite{barrett-cav19,barrett-neurips23}
to decide \eqref{eq:eaxp}, with $\varepsilon\in(0,1]$.

\begin{example} \label{ex:sexp}
  Consider the Pacman shortest path example in
  \autoref{fig:pacman}. Given a $\text{5} \times \text{5}$ grid map
  with cells either empty, actor, flag or ghost, it is relatively easy
  to compute the shortest path symbolically, although it may be
  challenging for a neural agent.
  \emph{Explaining} why the shortest path is at least some length
is a bit more complex, but not
  difficult.
  Here is an informal thought process applicable in this case.
  The shortest path always relies on the position of the actor and the
  flag (if they were closer to each other, this would make the path
  shorter).
  Then for determining the ghosts relevant for the shortest path, we
  can remove ghosts one by one: if the shortest path does not decrease
  in length, then the ghost we removed is not needed to explain the
  shortest path.
  %
  % We remove ghosts one by one, if the shortest path decreases, we
  % replace the ghost.
  %
  Once we have considered all ghosts, we have a minimal explanation of
  the shortest path.
  \autoref{fig:expl} depicts the grid cells required to explain why
  the shortest path is at least distance 5 as shown in \autoref{fig:grid}.
  \qed
\end{example}

\section{Hierarchical Explanations} \label{sec:app}

\begin{figure}[t]
  \input{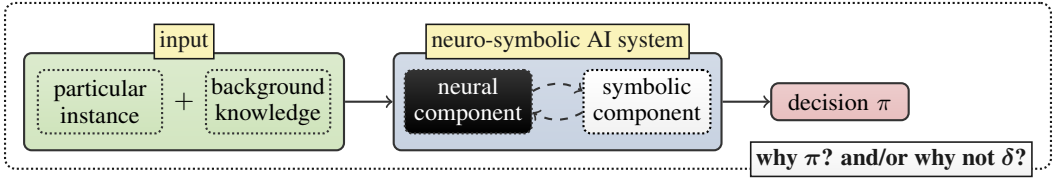}
  \caption{A general setup for explaining neuro-symbolic AI.}
  \label{fig:setup}
\end{figure}

Unlike purely neural AI systems, neuro-symbolic systems are meant to delegate
some responsibilities to their symbolic components.
This can negatively affect the model's interpretability.
First, while purely neural models are deemed an opaque black-box whose
decisions are hard for a human-decision maker to comprehend, the
problem is aggravated by the (complex) interactions between
the neural and symbolic components of a neuro-symbolic system.
Second, such interactions arguably
represent a challenge for post-hoc explanation approaches.
This applies to heuristic explainers, as they are known to be
susceptible to unsoundness issues~\cite{huang-corr23,jpms-corr23b}, and to
formal explanation approaches, as
their need  to deal with many
(typically NP-)hard computational problems often makes them become exponentially
harder with the increase of problem complexity.
The general setup for explaining a neuro-symbolic system can be seen
in~\autoref{fig:setup}.

Interestingly, the formal explainability of such hybrid neuro-symbolic
systems should in fact significantly benefit from this separation of
responsibilities.
First, explainability of the symbolic components is naturally achievable
by means of the advanced apparatus used by automated reasoning and discrete
optimization for dealing with over-constrained
systems~\cite{bs-dapl05,pms-aaai13,lm-cpaior13,iplms-cp15,lpmms-cj16}.
Namely, the symbolic rules can be seen as a set of constraints and,
given the inputs to the constraints passed by the neural component,
one can use the unsatisfiability of the constraints to identify an irreducible subset of the inputs responsible for
the symbolic decision.
As a result, one can solve this problem by applying mechanisms similar
to computing a (smallest) MUS of the set of rules, which are well
understood in discrete optimization~\cite[Ch.~21]{sat-handbook}.

Second, the functionality separation in neuro-symbolic AI leads
to the simplification of the neural component as it has to deal with
smaller or more focused sub-tasks, without a loss of performance of
the entire neuro-symbolic system. For instance, in the Pacman setting
of \autoref{fig:pacman}, the neural component serves to provide sensor
information classifying all the grid cells \emph{individually}.
This in turn may lead to structurally simpler neural models
and positively impacts the performance of formal reasoning engines on
such models. (Recall that scalability of formal explainability methods
is often seen as one of their limitations~\cite{msi-aaai22}.)

% \begin{figure}[b]
%   %
%   \centering
%   %
%   \begin{subfigure}[b]{0.298\textwidth}
%     %
%     \includegraphics[height=4.3cm]{example2a}
%     %
%     \caption{Symbolic explanation}
%     %
%     \label{fig:psymb}
%     %
%   \end{subfigure}%
%   %
%   \hspace{1.8cm}
%   %
%   \begin{subfigure}[b]{0.3\textwidth}
%     %
%     \includegraphics[height=4.3cm]{example2b}
%     %
%     \caption{Neural explanation}
%     %
%     \label{fig:pneur}
%     %
%   \end{subfigure}
%   %
%   \caption{Example o=f a hierarchical formal explanation for the Pacman puzzle considered in \autoref{fig:pacman}.}
%   %
% \label{fig:pexp}
%   %
% \end{figure}

The above two arguments provide us with the ground for the following
hierarchical approach to formal explanations for neuro-symbolic
systems.
Assume that we have a neuro-symbolic system whose setup implements the
pipeline of \autoref{fig:setup}, i.e., a neural model is trained to
provide sensor data to the symbolic reasoner that makes the final
decision.
(Note that such a setup is offered by
DeepProbLog~\cite{deraedt-aij21}, Scallop~\cite{naik-pldi23}, and
Pylon~\cite{ahmed-aaai22}.)
In this scenario, assuming there are $n\in\mbb{N}$ individual inputs
$\bar{\mbf{x}}=(\mbf{x}_1,\ldots,\mbf{x}_n)$, $\mbf{x}_i\in\mbb{F}$,
given to the system and, hence, the symbolic component receives $n$
individual (and independent of each other) neural decisions coming
from the neural component, the overall classification can be seen as
applying a composition of functions
$c=\kappa_\sigma\left(\kappa_\eta(\mbf{x}_1),\ldots,\kappa_\eta(\mbf{x}_n)\right)$
such that the neural component computes the function
$\kappa_\eta:\mbb{F}\rightarrow \mbb{Y}$ and its symbolic counterpart
implements $\kappa_\sigma:\prod_{j\in[n]}{\mbb{Y}}\rightarrow\fml{K}$.
Given a problem instance solved by such a system, an adaptation of the
concept of abductive explanation answering why a neuro-symbolic system
makes a certain decision can be obtained hierarchically by \emph{going
backward}.
First, the hierarchical explainer extracts an abductive explanation
for the symbolic decision $c\in\fml{K}$, which is provided as a subset
$\fml{Y}\subseteq\{1,\ldots,n\}$.
Then a formal abductive explanation~\cite{msi-aaai22} is extracted for
each of the inputs $\mbf{x}_j\in\mbb{F}$, $j\in\fml{Y}$,
independently of other inputs.
Formally, hierarchical abductive explanations can be defined as follows.
\begin{definition}[Hierarchical Abductive Explanation]
  \label{def:hexp}
  Let
  $c=\kappa_\sigma\left(\kappa_\eta(\mbf{x}_1),\ldots,\kappa_\eta(\mbf{x}_n)\right)$,
  $\mbf{x}_j\in\mbb{F}$, $j\in[n]$, be a decision made by a neuro-symbolic system
  involving a neural component $\kappa_\eta:\mbb{F}\rightarrow\mbb{Y}$
  and a symbolic component
  $\kappa_\sigma:\prod_{j\in[n]}{\mbb{Y}}\rightarrow\fml{K}$.
  Then a hierarchical abductive explanation for decision $c$ is a set
  $\fml{X}=\bigcup_{j\in\fml{Y}}{\fml{X}_j}$ such that
  $\fml{Y}\subseteq\{1,\ldots,n\}$ is an abductive explanation for the
  symbolic decision $c=\kappa_\sigma(y_1,\ldots,y_n)$ and each
  $\fml{X}_j$, $j\in\fml{Y}$, is an abductive explanation for the neural
  decision $y_j=\kappa_\eta(\mbf{x}_j)$.
\end{definition}
A concrete example of symbolic reasoning that can be used in the case
of the Pacman puzzle is provided in \autoref{ex:sexp}.
In general, this can be efficiently done using a formal reasoner
capable of handling the rules describing the symbolic component.
For example, one can use a generic explainer for ASP
programs~\cite{trieu-iclp23} or implement a bespoke explanation
extractor following the ideas behind MUS extraction widely used in the
analysis of over-constrained systems (see above).
(In our work, we encode the Datalog programs of Scallop into SAT and
apply an efficient smallest MUS extractor~\cite{iplms-cp15}.)
The result of this step is an irreducible set of neural inputs that
are deemed responsible for the symbolic decision.
Next, the explainer extracts an abductive explanation for
each of the individual parts of the neural inputs identified as
causing the decision in the first step.
Note that each such neural input is explained separately from the
other inputs, by applying the methodology of formal
XAI~\cite{msi-aaai22}.

\begin{figure}[t]
  \centering
  \begin{subfigure}[b]{0.24\textwidth}
    \includegraphics[width=\textwidth]{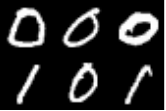}
    \caption{Input instance}
    \label{fig:linp}
  \end{subfigure}%
  \hfill
  \begin{subfigure}[b]{0.24\textwidth}
    \includegraphics[width=\textwidth]{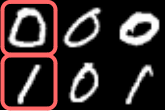}
    \caption{symbolic reason}
    \label{fig:lsymb}
  \end{subfigure}
  \hfill
  \begin{subfigure}[b]{0.24\textwidth}
    \includegraphics[width=\textwidth]{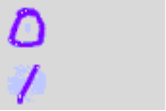}
    \caption{neural explanation}
    \label{fig:lneur}
  \end{subfigure}
  \caption{Hierarchical formal explanations. The system seeks to
  determine whether a given number handwritten in binary (top row) is
  greater than the other handwritten binary number (bottom row).}
\label{fig:lexp}
\end{figure}

\begin{example} \label{ex:hexp}
  Consider the problem of determining if a given handwritten binary 
  number is greater than another binary number.
  Let us focus on numbers of length 3.
  An example of such a problem instance is depicted in
  \autoref{fig:linp} and the prediction for this instance is
  \emph{false} because number 0 (represented as ``000'') is clearly
  not greater than number 5 (represented as ``101'').
  A neuro-symbolic model can be trained with a single neural component
  recognizing individual bits of both numbers (separately of each
  other) followed by a single symbolic rule
  \texttt{`greater(4*a + 2*b + 1*c > 4*x + 2*y + 1*z) :- d1(a), d2(b),
  d3(c), d4(x), d5(y), d6(z)'}.
  Here, each of the variables $\texttt{d}i\texttt{(}n\texttt{)}$
  represent the corresponding predictions made by the neural
  component.
  For instance, assuming the neural component is sufficiently
  accurate, we can conclude that digits $a$, $b$, and $c$ should be
  predicted as 0, 0, and 0 while digits $x$, $y$, and $z$ are to be
  predicted as 1, 0, and 1.
  Given this information, to explain why the first number is not
  greater than the second, we can consider a single (pseudo-Boolean)
  hard constraint $\fml{H}\triangleq(4a + 2b + c > 4x + 2y + z)$ and a
  set of soft clauses $\fml{S}\triangleq\{(\bar{a}), (\bar{b}),
  (\bar{c}), (x), (\bar{y}), (z)\}$.
  Running a smallest MUS enumerator~\cite{iplms-cp15} on formula
  $\fml{H}\land\fml{S}$ results in a symbolic reason for the
  prediction: $\{\bar{a},x\}$, shown in \autoref{fig:lsymb}.
  Next, we formally explain why digits $a$ and $x$ are predicted by
  the neural component as 0 and 1, respectively, using the apparatus
  of FXAI, i.e., see~\eqref{eq:axp} and~\eqref{eq:eaxp}.
	The resulting global abductive explanations are shown in
  \autoref{fig:lneur}.
  \qed
\end{example}

Our hierarchical setup for computing formal abductive explanations
requires all neural inputs to be independent of each other.
This is motivated by the fact that \emph{subset-minimal} hierarchical
abductive explanations can be obtained by following the above approach
with each of the two steps aiming for \emph{subset-minimal}
explanations $\fml{Y}$ and $\fml{X}_j$, $\forall j\in\fml{Y}$.
Indeed, it is not difficult to observe that the independence of the
inputs $\mbf{x}_j\in\mbb{F}$ implies that we can take the union of the
corresponding subset-minimal explanations $\fml{X}_j$ and
subset-minimality of the hierarchical explanation $\fml{X}$ will be
guaranteed.
Formally, this can be stated as follows.
\begin{proposition} \label{prop:min}
  Let
  $c=\kappa_\sigma\left(\kappa_\eta(\mbf{x}_1),\ldots,\kappa_\eta(\mbf{x}_n)\right)$,
  $\mbf{x}_j\in\mbb{F}$, $j\in[n]$, be a decision made by a neuro-symbolic system
  involving a neural component $\kappa_\eta:\mbb{F}\rightarrow\mbb{Y}$
  and a symbolic component
  $\kappa_\sigma:\prod_{j\in[n]}{\mbb{Y}}\rightarrow\fml{K}$.
  Also, let a set $\fml{X}=\bigcup_{j\in\fml{Y}}{\fml{X}_j}$ be a
  hierarchical abductive explanation for decision $c$ such that
  $\fml{Y}\in\{1,\ldots,n\}$ is an abductive explanation for the
  symbolic decision $c=\kappa_\sigma(y_1,\ldots,y_n)$ and each
  $\fml{X}_j$, $j\in\fml{Y}$, is an abductive explanation for the neural
  decision $y_j=\kappa_\eta(\mbf{x}_j)$.
  Then subset-minimality of the sets $\fml{Y}$ and $\fml{X}_j$ s.t.
  $j\in\fml{Y}$ guarantees that $\fml{X}$ is subset-minimal.
  \qed
\end{proposition}
Importantly, once the first step of the hierarchical approach finishes
by identifying the neural ``culprits'' of the decision made, one may
want to opt to apply a heuristic
explainer~\cite{guestrin-kdd16,lundberg-nips17,guestrin-aaai18} for
each (or some) of the individual neural inputs (instead of the formal
explainer), depending on user's needs.
Although this will affect the formal guarantees offered by FXAI
methods, such an alternative may often scale better in practice.
Similarly, instead of computing globally correct abductive
explanations~\eqref{eq:axp}, one can opt for utilizing an adversarial
robustness oracle and target local distance-restricted abductive
explanations with a particular distance $\varepsilon\in(0,1)$.
As a result, the choice of the formal explainer can be seen as a
trade-off between explanation soundness guarantees and the efficiency
of the overall approach.
Below we compare these options for several problems
considered.

% \anote{Here, we should cover:
%   \begin{itemize}
%     \item Explain the difficulties of this setting (interaction between components)
%     \item Argue that the symbolic part can be easily explained using the well-known apparatus of over-constrained systems.
%     \item Exemplify the above step (encoding + explanation computed) using the running example from \autoref{sec:prelim}.
%     \item Argue that this results in a subset of neural bits that need to be explained (independently of each other).
%     \item Continue the example showing how this works.
%     \item Discuss the use formal explanations for the neural part vs heuristic explanations and present it as a trade-off (their pros and cons: guarantees of soundness vs scalability)
%   \end{itemize}

\section{Experimental Results} \label{sec:res}

We use three  neuro-symbolic benchmarks
to evaluate the scalability and quality of the explanations produced by our hierarchical explanation framework in comparison to those of pure neural approaches with similar quality prediction results.
All experiments were conducted on a MacBook Pro with an Apple M3 Pro processor, with 12 CPU cores, 14 GPU cores, and 18 GB RAM.
(Source code and experimental data will be made publicly available with the final version of the paper.)

\subsection{Evaluated Explanation Methods}

As our hierarchical explanation framework can flexibly combine different explanation methods for neural components, we evaluate four configurations.
First, we consider the use of formal explanations for the neural components using Marabou~\cite{barrett-cav19} (dev. version 2024-03-11, BSD-3-Clause), a framework for verifying neural networks using SMT, and PySAT~\cite{imms-sat18} (v1.8.12, MIT).
Scallop is the language used to design the neuro-symbolic models (dev. version 2024-02-24, MIT).
Using this method, different values of $\varepsilon$ can be used requiring the explanation to hold for all compatible points in the $\varepsilon$-vicinity of the instance.
We configure Marabou with $\varepsilon = 1$, requiring the explanation to hold for all points, in configuration \texttt{HX-Marabou-e=1}.
The configuration \texttt{HX-Marabou-e<1} uses Marabou with $\varepsilon =
0.3$ (except for Pacman-SP where $\varepsilon =
0.2$).
For the third configuration \texttt{HX-SHAP}, we use our hierarchical framework in combination
with Shapley Additive Explanations (SHAP)~\cite{lundberg-nips17}  (v0.44.1, MIT) applied for explaining the neural component.
SHAP calculates the contribution of each feature to the model's prediction for a given input, identifying the most important features.
%
% Using these features a heuristic explanation is created.
%
% The configuration \texttt{HX-SHAP} uses SHAP as the explanation method for
% the neural component in the hierarchical setup.
% \pjs{SURELY HX-SHAP only uses SHAP for the neural component!!!!}
%
The next configuration \texttt{Scallop-SHAP}, directly builds (kernel) SHAP
explanations for the entire neuro-symbolic model treated as a black box.
The last configuration \texttt{NN-SHAP} uses baseline SHAP to obtain
explanations for purely neural solutions for each of the given benchmarks.
To achieve a similar accuracy for the purely neural applications, the size of the networks is too large for Marabou to create formal explanations within a reasonable timeframe.

\subsection{Benchmarks and Neural Architectures}

\begin{figure}[t]
  \centering
  \begin{subfigure}[b]{0.45\textwidth}
    \includegraphics[width=\textwidth]{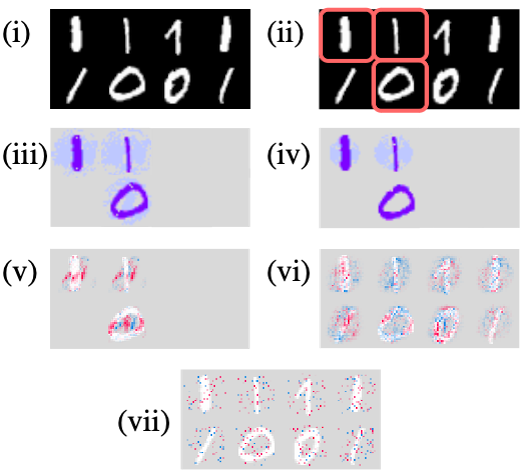}
    \caption{Positive example}
    \label{fig:lex8T}
  \end{subfigure}%
  \hfill
  \begin{subfigure}[b]{0.45\textwidth}
    \includegraphics[width=\textwidth]{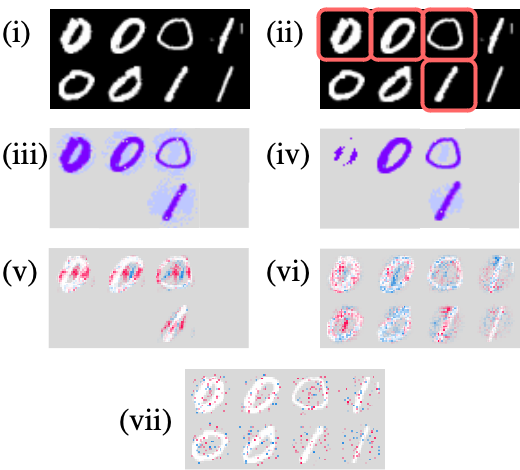}
    \caption{Negative example}
    \label{fig:lex8F}
    \end{subfigure}
  \caption{Given the two 4-digit binary numbers, explain whether the top number is greater than the bottom number.}
\label{fig:lex8}
\end{figure}

\begin{figure}[t]
  \centering
  \begin{subfigure}[b]{0.45\textwidth}
    \includegraphics[width=\textwidth]{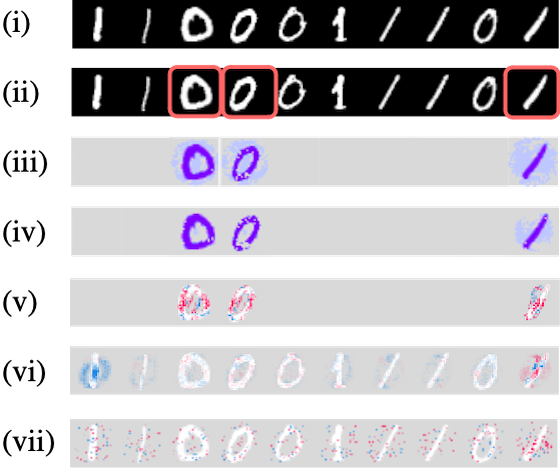}
    \caption{Positive example}
    \label{fig:rg10T}
  \end{subfigure}%
  \hfill
  \begin{subfigure}[b]{0.45\textwidth}
    \includegraphics[width=\textwidth]{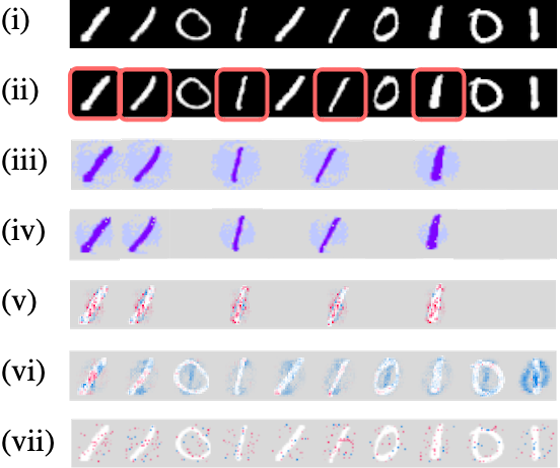}
    \caption{Negative example}
    \label{fig:rg10F}
    \end{subfigure}
  \caption{Explain whether the binary string is a member of the regular language $R_2$.}
\label{fig:regex10}
\end{figure}

\begin{figure}[t]
  \centering
  \begin{subfigure}[b]{0.23\textwidth}
    \includegraphics[width=\textwidth]{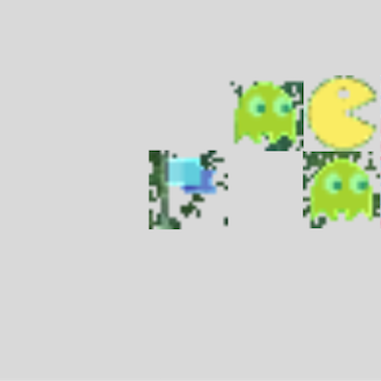}
    \caption{\texttt{HX-Marabou-e<1}}
    \label{fig:p4}
  \end{subfigure}
  \hfill
  \begin{subfigure}[b]{0.23\textwidth}
    \includegraphics[width=\textwidth]{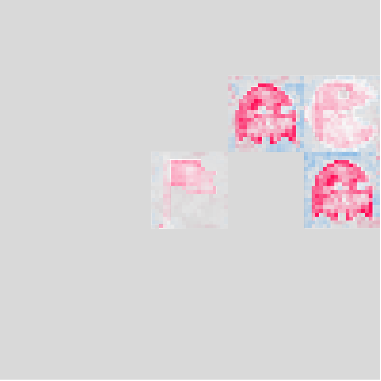}
    \caption{\texttt{HX-SHAP}}
    \label{fig:p5}
  \end{subfigure}
  \hfill
  \begin{subfigure}[b]{0.23\textwidth}
    \includegraphics[width=\textwidth]{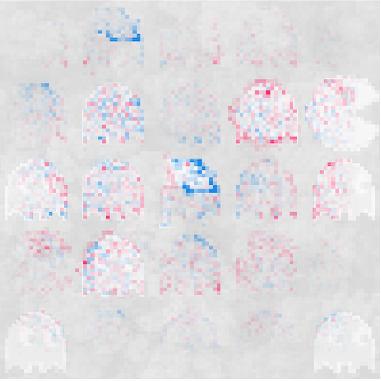}
    \caption{\texttt{NN-SHAP}}
    \label{fig:p6}
  \end{subfigure}
  \hfill
  \begin{subfigure}[b]{0.23\textwidth}
    \includegraphics[width=\textwidth]{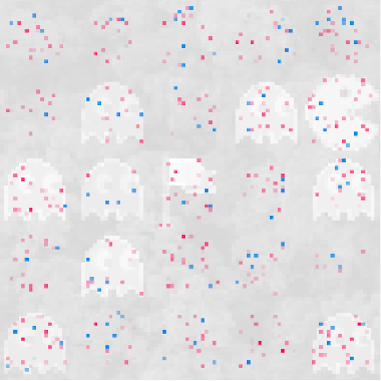}
    \caption{\texttt{Scallop-SHAP}}
    \label{fig:p7}
  \end{subfigure}
  \hfill
  \caption{[extending \autoref{fig:pacman}]. Explain why the length of the shortest path is at least 5.}
\label{fig:pac}
\end{figure}

% An example of each of
All the benchmarks are exemplified in \Crefrange{fig:lex8}{fig:pac}.
In \Crefrange{fig:lex8}{fig:regex10}, (i)~represents the input data, (ii)~the symbolic reduction, (iii)~the \texttt{HX-Marabou-e=1} explanation, (iv)~the \texttt{HX-Marabou-e<1} explanation, (v)~the \texttt{HX-SHAP} explanation, (vi)~the \texttt{Scallop-SHAP} explanation, and (vii)~the \texttt{NN-SHAP} explanation.
\autoref{fig:pac} shows the explanations (iv)--(vii) labelled. (See \autoref{fig:pacman} for the other parts.)

\paragraph*{Lex1-$\bm{n}$.}
Here we aim to learn a model that can predict the lexicographic order of two handwritten binary numbers, as discussed in \autoref{ex:hexp} and shown in \autoref{fig:lex8}.
All tasks take as input $n \in \{6,8\}$ images, randomly selected from the MNIST dataset~\cite{deng2012mnist} filtered to only contain the digits zero and one.
The combination of $n$ images in each task is unique.
The model is trained to predict whether the first number, represented by the first $\frac{n}{2}$ images, is lexicographically greater than the second number, represented by the remaining images.
In the neuro-symbolic system, a neural network predicts whether an image is zero or one, and a Scallop program then determines the lexicographic order based on the predicted values.
The neural model consists of 2 fully connected hidden layers of 10
nodes. Note that the neural component is only trained indirectly through Scallop.
For our pure neural application at $n=6$, we employ a network that uses 2 convolutional layers, followed by 4 fully connected hidden layers with 7500, 5000, 2500, and 64 nodes sequentially.
For $n=8$, we use 3 convolutional layers, but use only 2 fully connected hidden layers with 1024 and 256 nodes.

% %
% \begin{figure}[t]
%   %
%   \centering
%   %
%   \begin{subfigure}[b]{0.45\textwidth}
%     %
%     \includegraphics[width=\textwidth]{lex6T}
%     %
%     \caption{Positive example}
%     %
%     \label{fig:lex6T}
%     %
%   \end{subfigure}%
%   %
%   \hfill
%   %
%   \begin{subfigure}[b]{0.45\textwidth}
%     %
%     \includegraphics[width=\textwidth]{lex6F}
%     %
%     \caption{Negative example}
%     %
%     \label{fig:lex6F}
%     %
%     \end{subfigure}
%   %
%   \caption{Query: Is the top row 3-digit binary number greater than the bottom row 3-digit binary number. Answer: True or False}
%   %
% \label{fig:lex6}
%   %
% \end{figure}

\paragraph*{RegExp-$\bm{a}$-$\bm{n}$.}
Our regular expression benchmark, an example of which is shown in \autoref{fig:regex10}, aims to predict whether a string of 0-1 digits is in the language of regular expression $R_a$.
Each task takes an input of $n \in \{6,8,10\}$ images, randomly selected zero-one MNIST dataset similar to the Lex1-$n$ benchmark.
The model is then trained to recognize whether the string represented by the
images is in the language of the regular expression $R_1 =
\texttt{/1.*11.*0/}$, i.e., a string starting with a one, containing two
consecutive ones, and ending with a zero, or $R_2 =
\texttt{/(0.*11.*)|(.*00.*1)/}$, i.e., a string that either starts with a
zero and contains two consecutive ones or ends with a one and contains two consecutive zeroes.
We use $R_1$ together with the shorter string lengths $n \in \{6,8\}$ and $R_2$ with $n=10$.
Similar to Lex1-$n$ above, the neural component predicts the digit in the images, and
a Scallop program then determines whether the string is in the language of the regular expression based on the predicted values.
The pure neural application for RegExp-1-6 and RegExp-2-10 have layers identical to that of the Lex1-$6$ benchmark, with 7500, 5000, 1024, 10 nodes in the fully connected hidden layers sequentially for both, and that of RegExp-1-8 is the same as for Lex1-$8$, with 1024 and 10 nodes.
%
% \begin{figure}[t]
%   %
%   \centering
%   %
%   \begin{subfigure}[b]{0.35\textwidth}
%     %
%     \includegraphics[width=\textwidth]{rg6T}
%     %
%     \caption{Positive example}
%     %
%     \label{fig:rg6T}
%     %
%   \end{subfigure}%
%   %
%   \hfill
%   %
%   \begin{subfigure}[b]{0.35\textwidth}
%     %
%     \includegraphics[width=\textwidth]{rg6F}
%     %
%     \caption{Negative example}
%     %
%     \label{fig:rg6F}
%     %
%     \end{subfigure}
%   %
%   \caption{Query: This 6-digit binary string is a member of the regular language, $L = 1(0\mid1)^*11(0\mid1)^*0$? Answer: True or False}
%   %
% \label{fig:regex6}
%   %
% \end{figure}
% %
% \begin{figure}[t]
%   %
%   \centering
%   %
%   \begin{subfigure}[b]{0.4\textwidth}
%     %
%     \includegraphics[width=\textwidth]{rg8T}
%     %
%     \caption{Positive example}
%     %
%     \label{fig:rg8T}
%     %
%   \end{subfigure}%
%   %
%   \hfill
%   %
%   \begin{subfigure}[b]{0.4\textwidth}
%     %
%     \includegraphics[width=\textwidth]{rg8F}
%     %
%     \caption{Negative example}
%     %
%     \label{fig:rg8F}
%     %
%     \end{subfigure}
%   %
%   \caption{Query: This 8-digit binary string is a member of the regular language, $L = 1(0\mid1)^*11(0\mid1)^*0$? Answer: True or False}
%   %
% \label{fig:regex8}
%   %
% \end{figure}
%

\paragraph*{Pacman-SP.}
As discussed in \Cref{ex:rex,ex:sexp} and shown in \autoref{fig:pac}, the aim of this benchmark is to find the length of a shortest grid path from the actor's start position to a target flag position without stepping onto squares containing ghosts.
Using the same images and obfuscation technique as~\cite{naik-pldi23} (MIT), we randomly generate grids of size $5 \times 5$ containing the actor, a target flag, and eight ghost obstacles each in a different position.
Our neuro-symbolic application uses a neural network, consisting of a single fully connected layer of size 128, to predicts whether a cell on the grid is empty, contains the actor, the flag, or a ghost.
A Scallop program then determines the shortest path based on the predicted values.
The pure neural approach uses a network with 4 convolutional layers and 3 fully connected hidden layers with 4096, 2048, and 64 nodes, respectively.

\ignore{
\paragraph*{General comments.}
The CNN architectures above are in general much larger than the
neuro-symbolic models used.
Also, similar to the previous results in \cite{naik-pldi23}, we find
that the neuro-symbolic models take significantly less time to train,
achieving higher test accuracy earlier.
}
\subsection{Evaluation}
\paragraph{Model Training Time Statistics.}
\begin{table*}[h!]
  \setlength{\tabcolsep}{4pt}
  \caption{\label{tab:train}Training time and prediction accuracy of the neuro-symbolic and purely neural models. For each model the columns depict the total training time (MM:SS), number of training epochs, time per epoch (MM:SS), final test accuracy (20\% of data), and final training accuracy (80\% of data).}
  % \begin{adjustbox}{center}
  \resizebox{\textwidth}{!}{
  \begin{tabular}{@{}lrrrrrrrrrr@{}}
    \toprule
& \multicolumn{5}{c}{Scallop (Neuro-Symbolic)} & \multicolumn{5}{c}{Convolutional Neural Network (CNN)} \\
\cmidrule(l){2-6}\cmidrule(l){7-11}
Benchmark   & Time    & E. & E. Time & Test (\%) & Train (\%) & Time    & E. & E. Time & Test (\%) & Train (\%) \\
\midrule
Lex1-6      & 00:12 & 3      & 00:04 & 99.90          & 95.00           & 02:42 & 3      & 00:54 & 98.54         & 96.38          \\
Lex1-8      & 00:12 & 2      & 00:06 & 99.92          & 94.09           & 02:12 & 4      & 00:34 & 96.78         & 95.88          \\
RegExp-1-6  & 00:20 & 5      & 00:04 & 99.50          & 94.45           & 01:43 & 2      & 00:51 & 99.14         & 98.18          \\
RegExp-1-8  & 00:15 & 3      & 00:05 & 99.54          & 93.64           & 02:12 & 4      & 00:33 & 95.94         & 94.38          \\
RegExp-2-10 & 00:24 & 3      & 00:08 & 99.78          & 94.29           & 05:04 & 3      & 01:41 & 96.70         & 96.54          \\
Pacman-SP   & 19:36 & 3      & 06:32 & 100.00         & 99.90           & 31:29 & 12     & 02:37 & 89.15         & 95.98          \\
\bottomrule
\end{tabular}
% \end{adjustbox}
}
\end{table*}

\autoref{tab:train} shows the time required for both the Scallop and CNN models to get trained to achieve a high accuracy.
%
%Similar to previous results in \cite{naik-pldi23}, we find that the neuro-symbolic models using Scallop achieve higher accuracy earlier.
%
%This is also evident as size of the architectures of the CNN models are much larger to perform the same tasks.
%
%This results in a significant time difference to achieve better accuracy than CNN models.
%
%Each of the models achieves the mentioned accuracy during training and testing in the given number of epochs, where each epoch time and the total time for training are mentioned in the table for all the benchmarks considered.
%
In the specified number of epochs, these models obtain the mentioned accuracies for their respective benchmarks as reported in the table.
Note that the epoch time in table is the time taken for training the corresponding model in a single epoch.
\autoref{tab:train} indicates that there is a significant training time advantage for neuro-symbolic models over CNN models to achieve high accuracy.
This is also evident as size of the architectures of the CNN models (detailed in the previous subsections) are much larger to perform the same tasks, as established in \cite{naik-pldi23}.

\paragraph{On Explanation Sizes.}
\begin{table*}[h!]
  \setlength{\tabcolsep}{3pt}
  \caption{\label{tab:expl} Explanation statistics for each evaluated explanation methods and benchmarks. Columns min, avg, max, and time depict the minimum, average, and maximum explanation size, and average explanation time (MM:SS or HH:MM:SS), respectively.}
  % \begin{adjustbox}{center}
  \resizebox{\textwidth}{!}{
  \begin{tabular}{@{}lrrrrrrrrrrrrrrrrrrrr@{}}
    \toprule
    & \multicolumn{4}{c}{ \texttt{HX-Marabou-e=1} } & \multicolumn{4}{c}{ \texttt{HX-Marabou-e<1} } & \multicolumn{4}{c}{ \texttt{HX-SHAP} } & \multicolumn{4}{c}{ \texttt{Scallop-SHAP} } & \multicolumn{4}{c}{ \texttt{NN-SHAP} } \\
    \cmidrule(l){2-5}\cmidrule(l){6-9}\cmidrule(l){10-13}\cmidrule(l){14-17}\cmidrule(l){18-21}
    Benchmark & min & avg & max & time & min & avg & max & time & min & avg & max & time & min & avg & max & time & min & avg & max & time \\
    \midrule
    Lex1-6 & 1027 & 1413 & 2388 & 02:04 & 576 & 932 & 1648 & 01:54 & 894 & 1131 & 1790 & 00:01 & 138 & 430 & 488 & 08:55 & 2641 & 2687 & 2753 & 00:07\\
    Lex1-8 & 733 & 1130 & 2116 & 02:25 & 302 & 607 & 1188 & 02:10 & 910 & 1242 & 2276 & 00:01 & 316 & 459 & 498 & 21:25 & 3480 & 3558 & 3612 & 00:04\\
    RegExp-1-6 & 326 & 571 & 1854 & 00:50 & 70 & 328 & 1185 & 00:48 & 470 & 705 & 2029 & 00:01 & 233 & 406 & 492 & 08:31 & 2616 & 2679 & 2742 & 00:08\\
    RegExp-1-8 & 496 & 885 & 2377 & 01:03 & 273 & 633 & 1866 & 01:26 & 876 & 1196 & 2190 & 00:01 & 297 & 435 & 490 & 16:37 & 3486 & 3550 & 3626 & 00:08\\
    RegExp-2-10 & 832 & 1422 & 2778 & 02:42 & 296 & 742 & 1522 & 02:18 & 972 & 1570 & 3126 & 00:01 & 191 & 435 & 498 & 25:53 & 4408 & 4474 & 4579 & 00:10\\
    Pacman-SP & 721 & 1617 & 2902 & 08:13 & 398 & 1033 & 1784 & 10:04 & 800 & 1728 & 3200 & 00:01 & 391 & 456 & 522 & 20:45:32 & 6957 & 7038 & 7040 & 05:14\\
    \bottomrule
  \end{tabular}
  % \end{adjustbox}
}
\end{table*}

The sizes of the explanations and the average time to compute them is
detailed in \autoref{tab:expl}.
The data demonstrates that the explanation size from
\texttt{HX-Marabou-e=1} is larger than that from
\texttt{HX-Marabou-e<1} (which is to be expected) and comparable to
that from \texttt{HX-SHAP}.
However, \texttt{HX-Marabou-e=1} explanations are global abductive
explanations and guaranteed to be correct in the entire feature space,
i.e., if these input pixels are left unchanged the result will always
be the same regardless of other inputs.

The average explanation sizes of \texttt{HX-Marabou-e=1}, \texttt{HX-Marabou-e<1}, \texttt{HX-SHAP}, \texttt{Scallop-SHAP} and \texttt{NN-SHAP} are around $16\%$, $9\%$, $18\%$, $7\%$ and $60\%$, respectively, of the total number of features in the instance.
This not only indicates that it is challenging to explain
(convolutional) neural networks as large as ours but also proves the
efficacy of neuro-symbolic models.
Consider \autoref{fig:rg10T}, the \texttt{NN-SHAP} explanations
focus more on the first and last digits for the
prediction, whereas common sense dictates the first digit to be
irrelevant to the prediction (given the regular expression).
The \texttt{Scallop-SHAP} explanations are also scattered around all the digits in the image, not producing any effective explanation for the prediction.
Similar observations can be made w.r.t. the other examples shown, demonstrating that \texttt{Scallop-SHAP} and \texttt{NN-SHAP} essentially get lost while trying to explain the decision.
Also, one can rely on explanations generated by \texttt{HX-SHAP} if time is a
constraint, but at the cost of formal correctness.
%%%%%%%%%%%%%%%%%%%%%%%%%%%%%%%%%%%%%%%%%%%%%%%%%%%%%%%%%%%%

\paragraph{On Explanation Quality.}
\begin{table*}[h!]
  \setlength{\tabcolsep}{3pt}
  \caption{\label{tab:explPercent} Percentage of neural components and explanation size for each of the evaluated explanation methods and benchmarks. Columns min, avg, and max depict percentage ($\%$) of the minimum, average, and maximum number of neural inputs in the first row and that of explanation size in the row below, respectively.}
  % \begin{adjustbox}{center}
  \resizebox{\textwidth}{!}{
  \begin{tabular}{@{}lrrrrrrrrrrrrrrr@{}}
    \toprule
    & \multicolumn{3}{c}{ \texttt{HX-Marabou-e=1} } & \multicolumn{3}{c}{ \texttt{HX-Marabou-e<1} } & \multicolumn{3}{c}{ \texttt{HX-SHAP} } & \multicolumn{3}{c}{ \texttt{Scallop-SHAP} } & \multicolumn{3}{c}{ \texttt{NN-SHAP} } \\
    \cmidrule(l){2-4}\cmidrule(l){5-7}\cmidrule(l){8-10}\cmidrule(l){11-13}\cmidrule(l){14-16}
    Benchmark & min & avg & max & min & avg & max & min & avg & max & min & avg & max & min & avg & max \\
    \midrule
	  \multirow{2}{*}{Lex1-6} & 33.33 & 41.83 & 66.67 & 33.33 & 41.83 & 66.67 & 33.33 & 41.83 & 66.67 & 100 & 100 & 100 & 100 & 100 & 100\\
	  & 15.86 & 23.10 & 37.58 & 6.33 & 11.73 & 19.54 & 16.60 & 23.55 & 37.50 & 2.93 & 9.14 & 10.37 & 56.14 & 57.12 & 58.52\\
	  \midrule
	  \multirow{2}{*}{Lex1-8} & 25.00 & 34.12 & 62.50 & 25.00 & 34.12 & 62.50 & 25.00 & 34.12 & 62.5 & 100 & 100 & 100 & 100 & 100 & 100\\
	   & 11.69 & 17.95 & 34.36 & 4.82 & 9.64 & 19.15 & 16.07 & 22.54 & 42.17 & 5.04 & 7.31 & 7.94 & 55.48 & 56.72 & 57.58\\
	  \midrule
	  \multirow{2}{*}{RegExp-1-6} & 16.67 & 23.17 & 66.67 & 16.67 & 23.17 & 66.67 & 16.67 & 23.17 & 66.67 & 100 & 100 & 100 & 100 & 100 & 100\\
	   & 6.93 & 12.14 & 39.41 & 1.49 & 6.97 & 25.19 & 9.99 & 14.99 & 43.13 & 4.95 & 8.63 & 10.45 & 55.61 & 56.95 & 58.29\\
	  \midrule
	  \multirow{2}{*}{RegExp-1-8} & 12.50 & 16.75 & 50.00 & 12.50 & 16.75 & 50.00 & 12.50 & 16.75 & 50.00 & 100 & 100 & 100 & 100 & 100 & 100\\
	   & 7.95 & 12.29 & 37.57 & 4.86 & 8.96 & 28.82 & 7.99 & 11.00 & 32.97 & 4.74 & 6.94 & 7.81 & 55.58 & 56.60 & 57.81\\
	  \midrule
	  \multirow{2}{*}{RegExp-2-10} & 20.00 & 30.90 & 60.00 & 20.00 & 30.90 & 60.00 & 20.00 & 30.90 & 60.00 & 100 & 100 & 100 & 100 & 100 & 100\\
	   & 10.61 & 18.13 & 35.43 & 3.78 & 9.46 & 19.41 & 12.40 & 20.02 & 39.87 & 2.44 & 5.55 & 6.35 & 56.22 & 57.06 & 58.40\\
	  \midrule
	  \multirow{2}{*}{Pacman-SP} & 8.00 & 17.76 & 32.00 & 8.00 & 17.76 & 32.00 & 8.00 & 17.76 & 32.00 &100 & 100 & 100 & 100 & 100 & 100\\
	   & 7.21 & 16.17 & 29.02 & 3.98 & 10.33 & 17.84 & 8.00 & 17.28 & 32.00 & 3.91 & 4.56 & 5.22 & 69.57 & 70.38 & 70.40\\
    \bottomrule
  \end{tabular}
  % \end{adjustbox}
}
\end{table*}
\autoref{tab:explPercent} gives insights on the percentage of features contributing to explanations.
For each benchmark, the first row presents the percentage of the minimum, average and maximum number of neural inputs considered in the explanations generated through the configurations \texttt{HX-Marabou-e=1}, \texttt{HX-Marabou-e<1}, \texttt{HX-SHAP}, \texttt{Scallop-SHAP}, and \texttt{NN-SHAP}.
The second row of each benchmark presents the percentage of the minimum, average and maximum size of explanations, generated in all of the configurations.

Since the first three configurations follow the hierarchical symbolic reasoning, the number (and thus, the percentage) of neural inputs responsible for the explanation is same.
On the contrary, explanations from the \texttt{Scallop-SHAP} and \texttt{NN-SHAP} configurations are distributed over all of the neural inputs.
However, the percentage of explanation size of \texttt{Scallop-SHAP}, where SHAP explains the Scallop model as a blackbox, is the smallest but the explanation makes no sense.
Similarly for the \texttt{NN-SHAP} configuration, the explanation size is nearly one-third of the total feature space, substantially losing its essence of explaining the model's prediction.
Observe that the percentage of average explanation size of \texttt{HX-Marabou-e=1} is comparable, but lower than that of \texttt{HX-SHAP} (except for RegExp-1-8).

Consider the percentage of average explanation sizes for the Pacman-SP benchmark.
The \texttt{HX-Marabou-e=1} explanations that guarantees formal correctness is merely $16.17\%$.
For the \texttt{HX-Marabou} configuration at $\varepsilon=0.2$, the percentage is $10.33$.
For the \texttt{HX-SHAP} and \texttt{Scallop-SHAP} configurations, the percentage is $17.28$ and $4.56$ respectively, and that of \texttt{NN-SHAP}, it is a whopping $70.38\%$ of the entire feature space.
%%%%%%%%%%%%%%%%%%%%%%%%%%%%%%%%%%%%%%%%%%%%%%%%%%%%%%%%%%%%

\paragraph{Sorting Heuristics for \texttt{\textbf{HX-Marabou}}.}
The quality of abductive explanations generated by \texttt{HX-Marabou} depends on the feature traversal procedure undertaken.
Consider an image of size $28\times28$ as an input, which is to be explained.
If the traversal begins from the top left of the image towards the bottom right, the features on the top of the image will be eliminated first and thus, a majority of the features in an explanation will lie towards the bottom of the image.
Similarly, if the traversal is initiated from the top right corner towards the bottom left, majority of the features in an explanation will lie towards the left of the image.
This creates a need to apply heuristics such that our model eliminates the features that surely do not contribute to the prediction and then focusses on the important features.

In our setup, we iterate over the pixels/features based on distance from the centre of the image (of the individual neural inputs), and saturation and lightness of the pixels.
The distance measure ensures that we first eliminate the features that reside close to the boundaries and the corners of the image.
The saturation and lightness of pixels are considered to eliminate the darker ones first.
For grayscale images of benchmarks Lex1-$\bm{n}$ and RegExp-$\bm{a}$-$\bm{n}$, the saturation and lightness measure is equivalent to the brightness of the pixel (measured by the value of the pixel: the higher, the brighter).
The saturation and lightness is a reasonable measure since our benchmarks of digits and pacman grid cells have bright and/or coloured pixels that are responsible for the prediction.
On applying these heuristics, the abductive explanations generated are of improved quality and give better visual representation.

\section{Related Work} \label{sec:relw}

There are a large variety of approaches of
XAI~\cite{monroe-cacm18,lipton-cacm18,miller-aij19}, including
interpretable model
synthesis~\cite{leskovec-kdd16,rudin-kdd17a,molnar-bk20} and post-hoc
explainability of black-box ML
models~\cite{guestrin-kdd16,lundberg-nips17,guestrin-aaai18,miller-aij19}.
The approach to explainability closest to ours is based on formal
reasoning about the model of interest.
It seeks to compute so-called
formal abductive and contrastive explanations using a series of oracle
calls to a reasoning engine and proving certain properties of the
model of interest~\cite{darwiche-ijcai18,inms-aaai19,msi-aaai22}.
The area of FXAI is tightly related~\cite{inms-nips19} to adversarial
robustness checking and neural model
verification~\cite{kwiatkowska-ijcai18,lomuscio-kr18,barrett-cav19};
hence, the reasoning engines developed in the latter community can be
applied for computing formal
explanations~\cite{barrett-neurips23,huang-corr23b}.

Neuro-symbolic AI is often seen as a response to the weaknesses of
purely neural ML architectures and often deemed crucial for
constructing rich computational cognitive models where reasoning is
required~\cite{valiant-cacm84,dillig-neurips20,russo-ml22,garcez-air23}.
A modern generation of neuro-symbolic systems implements ways to integrate
conventional neural model training with the power of existing symbolic reasoning
methods to solve complex problems.
These systems include DeepProbLog~\cite{deraedt-aij21}, Pylon~\cite{ahmed-aaai22}, and
Scallop~\cite{naik-pldi23}, which utilize Prolog, constraint modelling, and Datalog engines respectively.
Due to the use of symbolic reasoning, neuro-symbolic AI is often
deemed more trustable, interpretable than neural AI and also
self-explanatory~\cite{angelov-wire21,subbarao-aaai22,piplai-ic23,omicini-fedcsis23}.
We are not aware of other approaches to (formal) post-hoc
explainability of neuro-symbolic AI.

% \anote{Topics to cover:
%   \begin{itemize}
%     \item XAI and FXAI
%     \item neuro-symbolic systems (we use Scallop but there are other systems)
%     \item explanations for neuro-symbolic AI (if any!)
%   \end{itemize}
% }

\section{Conclusions} \label{sec:conc}

This paper proposes a formal approach to computing abductive
explanations for neuro-symbolic AI systems.
The approach applies in the case of neuro-symbolic systems whose
inputs are independent of each other and, as a result, implements the
extraction of abductive explanations \emph{hierarchically}.
Importantly, given subset-minimal abductive explanations for the individual
decisions made by the neural and symbolic components of the system,
the approach is guaranteed to report a subset-minimal abductive
explanation for the entire system.
Experimental results obtained on a few families of benchmarks
demonstrate the high quality of hierarchical explanations.
A few lines of future work can be envisioned.
First, it is interesting to investigate how the relaxation of the
input independence requirement impacts the quality of neuro-symbolic
explanations and the difficulty of their extraction.
Second, while input independence facilitates our approach to the
computation of subset-minimal AXps, it does not directly simplify the
process of computing (i)~contrastive explanations (CXps) and
(ii)~cardinality-minimal AXps, which we are planning to address next.
%
% A natural line of future work is to
% investigate how the relaxation of the input independence requirement
% impacts the quality of neuro-symbolic explanations and the difficulty
% of their extraction.

% \anote{What we did and what lines of future work we envision.}

% \begin{ack}
% This research was (fully/ partially) funded by the Australian Government through the Australian Research Council Industrial Transformation Training Centre in Optimisation Technologies, Integrated Methodologies, and Applications (OPTIMA), Project ID IC200100009.
% \end{ack}

\printbibliography

%%%%%%%%%%%%%%%%%%%%%%%%%%%%%%%%%%%%%%%%%%%%%%%%%%%%%%%%%%%%

%\input{appendix}

\end{document}